\documentclass[letterpaper]{article} 
\usepackage{aaai2026}  
\usepackage{times}  
\usepackage{helvet}  
\usepackage{courier}  
\usepackage[hyphens]{url}  
\usepackage{graphicx} 
\usepackage{natbib}  
\usepackage{caption} 
\usepackage{afterpage} 
\usepackage{amsmath}
\usepackage{amssymb}
\usepackage{amsfonts}
\usepackage{booktabs}      
\usepackage{multirow}      
\usepackage{tikz}
\usetikzlibrary{external}
\usepackage{algorithm}
\usepackage{algorithmic}
\usepackage{caption}
\usepackage{newfloat}
\usepackage{listings}
\DeclareCaptionStyle{ruled}{labelfont=normalfont,labelsep=colon,strut=off} 
\floatstyle{ruled}
\newfloat{listing}{tb}{lst}{}
\floatname{listing}{Listing}
\title{X-MoGen: Unified Motion Generation across Humans and Animals}

\author{
Xuan Wang\textsuperscript{\rm 1}\equalcontrib \quad
Kai Ruan\textsuperscript{\rm 3}\equalcontrib \quad
Liyang Qian\textsuperscript{\rm 1} \quad
Zhizhi Guo\textsuperscript{\rm 2} \quad
Chang Su\textsuperscript{\rm 1} \quad
Gaoang Wang\textsuperscript{\rm 1\thanks{Corresponding author.}}
}
\affiliations{
\textsuperscript{\rm 1} Zhejiang University \quad
\textsuperscript{\rm 2} Institute of Artificial Intelligence (TeleAI), China Telecom \quad
\textsuperscript{\rm 3} Renmin University of China \\[6pt]

\texttt{xuanw@zju.edu.cn}, 
    \texttt{kairuan@ruc.edu.cn}, 
    \texttt{gaoangwang@intl.zju.edu.cn}
}

\usepackage{bibentry}

\begin{document}

\maketitle

\begin{abstract}
    
Text-driven motion generation has attracted increasing attention due to its broad applications in virtual reality, animation, and robotics. While existing methods typically model human and animal motion separately, a joint cross-species approach offers key advantages, such as a unified representation and improved generalization. However, morphological differences across species remain a key challenge, often compromising motion plausibility. To address this, we propose \textbf{X-MoGen}, the first unified framework for cross-species text-driven motion generation covering both humans and animals. X-MoGen adopts a two-stage architecture. First, a conditional graph variational autoencoder learns canonical T-pose priors, while an autoencoder encodes motion into a shared latent space regularized by morphological loss. In the second stage, we perform masked motion modeling to generate motion embeddings conditioned on textual descriptions. During training, a morphological consistency module is employed to promote skeletal plausibility across species. To support unified modeling, we construct \textbf{UniMo4D}, a large-scale dataset of 115 species and 119k motion sequences, which integrates human and animal motions under a shared skeletal topology for joint training. Extensive experiments on UniMo4D demonstrate that X-MoGen outperforms state-of-the-art methods on both seen and unseen species.

\end{abstract}

\begin{figure}[h]
    \centering
    \includegraphics[width=\columnwidth]{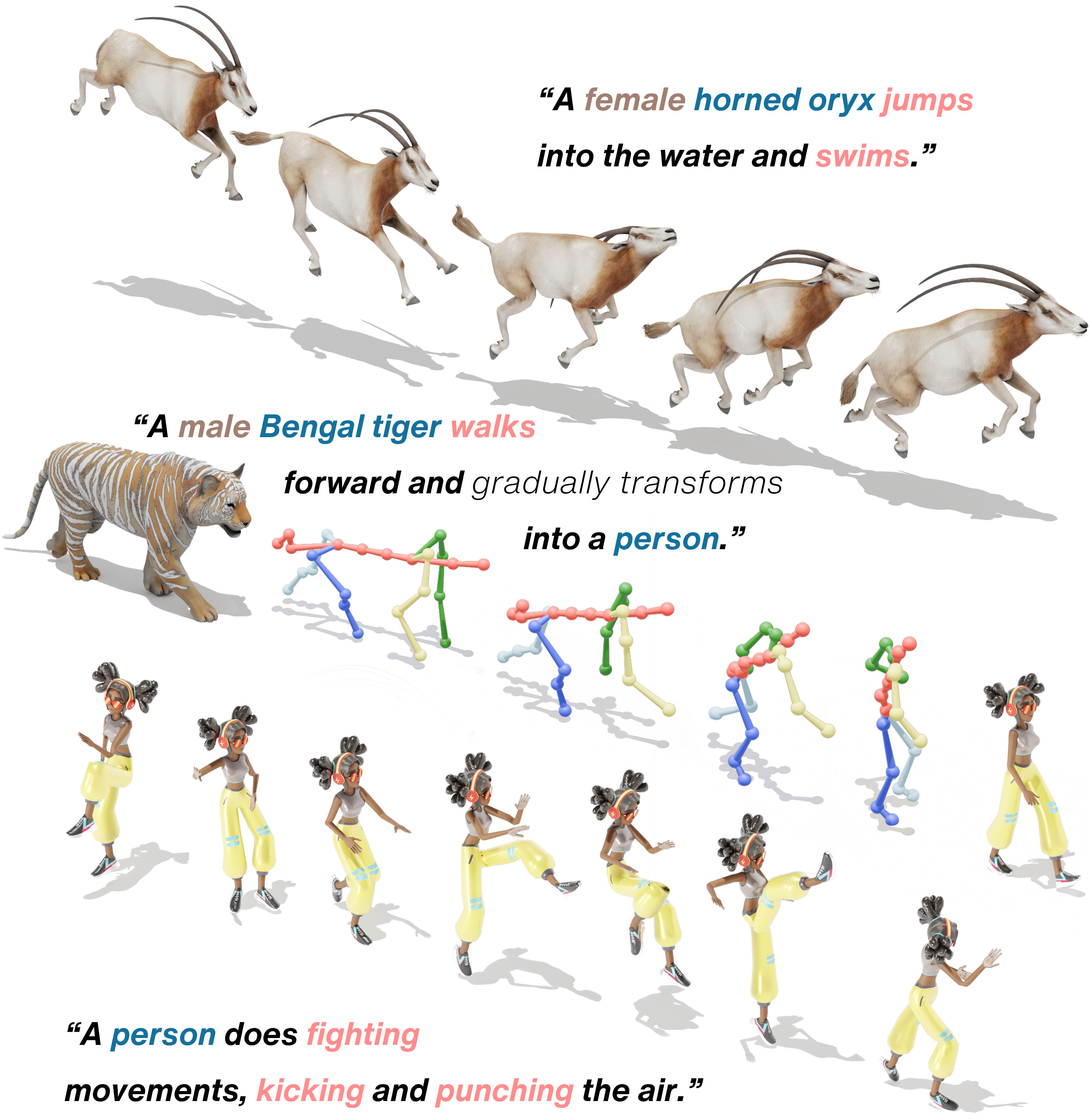}
    \caption{X-MoGen achieves a wide range of capabilities within a single unified framework, including generating both human and animal motions from text descriptions and enabling smooth cross-species motion transitions.}
    \label{fig:teaser}
\end{figure}

\section{Introduction}

Motion generation aims to synthesize high-quality, physically plausible 3D motion sequences. It has become an active research area with wide-ranging applications in film production, video games, virtual/augmented reality, and robotics~\cite{zhu2023human,sahili2025text}. Various control modalities have been explored for motion generation, including text~\cite{t2m}, music~\cite{siyao2022bailando}, sketch-based input~\cite{zhong2025sketch2anim}, and multimodal signals~\cite{guo2025motionlab,zhang2024large}. Of these modalities, text has emerged as a particularly popular interface due to its intuitive and expressive nature. Despite recent advances, most text-driven motion generation methods focus exclusively on either human~\cite{t2m,hong2025salad} or animal~\cite{wang2025animo} motion, treating them as independent tasks. However, unified modeling offers two key advantages~\cite{yang2024x}: (1) Unified representation learning, where aligning motion semantics across species leads to more structured and expressive latent spaces; and (2) Cross-species generalization, as diverse body morphologies and motion patterns improve transferability to unseen species and novel actions.

Previous works typically struggle to achieve unified modeling due to incompatible skeletal structures across datasets, necessitating separate models for each dataset~\cite{guo2024momask, hong2025salad}. To address this, we introduce \textbf{UniMo4D}, a large-scale dataset covering humans and 114 animal species. UniMo4D standardizes skeletal topology while retaining species-specific morphology, facilitating learning in a shared representation space. This provides a foundation for joint human-animal motion modeling. However, unifying such diverse species introduces new challenges, as variations in morphology and motion dynamics make it difficult to generate consistent and realistic cross-species motions with a single model.

To address these challenges, we propose \textbf{X-MoGen}, a unified framework designed for human-animal motion generation. As illustrated in Figure~\ref{fig:teaser}, X-MoGen synthesizes motions for both humans and animals from a single text description, providing flexible and generalizable control. It also supports applications such as cross-species transformation by producing smooth and semantically consistent transitions, such as transforming a tiger into a human.
X-MoGen operates in two stages. The first stage uses a conditional graph variational autoencoder to generate species-specific canonical T-pose priors, while a motion autoencoder encodes motion sequences into a compact latent space regularized by a morphological loss. In the second stage, motion is synthesized using a masked modeling mechanism~\cite{li2023mage,li2024autoregressive} conditioned on the input text. To maintain structural coherence, a morphological consistency module predicts skeletal lengths from the latent representation and incorporates them into the loss computation.

We evaluate X-MoGen on the UniMo4D dataset against several state-of-the-art baselines. Results show that our model achieves strong performance on both seen and unseen species, demonstrating its ability to generalize across diverse body morphologies. Quantitative results further confirm that X-MoGen generates realistic motions that are consistent with species-specific morphology and dynamics.

Our main contributions are summarized as follows:
\begin{itemize}
    \item To the best of our knowledge, \textbf{X-MoGen} is the first model designed for text-driven motion generation across humans and animals, addressing challenges in morphological variation and cross-species generalization.
    
    \item We construct \textbf{UniMo4D},  a large-scale benchmark that combines human and animal motion data with a unified skeletal topology, enabling joint training across species.

    \item Extensive experiments demonstrate that our method achieves strong performance on both seen and unseen species, supported by both qualitative and quantitative evaluations.
\end{itemize}

\section{Related Work}

\subsection{Text-Driven Human Motion Generation}

Text-driven motion generation methods can be broadly categorized into quantization-based and non-quantization-based approaches. Quantization-based methods~\cite{jiang2023motiongpt,zhang2023generating,guo2024momask,zhong2023attt2m,pinyoanuntapong2024mmm,liao2025shape} discretize motion sequences using various quantization schemes and predict motion tokens from text. However, this discretization often leads to information loss.
Alternatively, non-quantization-based approaches operate directly in the continuous motion space~\cite{rombach2022high,tevet2023human,zhang2023finemogen,zhang2024motiondiffuse,meng2025rethinking,hong2025salad}. For example, MDM~\cite{tevet2023human} employs a diffusion model on raw motion sequences for synthesis. Subsequent works leverage latent diffusion models~\cite{rombach2022high} to improve generation quality while reducing computational cost. MARDM~\cite{meng2025rethinking} introduces a masked autoregressive framework for high-fidelity motion generation. However, such models are typically trained on skeletons with fixed bone lengths. The recent ShapeMove~\cite{liao2025shape} supports variable bone lengths by learning additional shape features, allowing skeletal morphology to be controlled by text. In contrast, X-MoGen learns species-specific motions without relying on additional shape features, directly adapting to skeletal variations present in the data.

\subsection{Text-Driven Animal Motion Generation}

Generating animal motions from text is inherently more complex than human motion generation, due to the vast diversity in animal morphology and behavior. OmniMotionGPT~\cite{yang2024omnimotiongpt} models both human and animal motions, enabling animal synthesis via transfer learning from human motion–text pairs. Crucially, human motion serves as a key condition, guiding animal motion generation within the same category. Motion Avatar~\cite{zhang2024motion} employs large language models as high-level planners to generate 3D animal avatar meshes. AniMo~\cite{wang2025animo} introduces spatiotemporal modeling and species-aware modulation for text-driven animal motion generation. 
However, these methods do not explicitly address morphological variations across species. In contrast, our approach incorporates T-pose encoding and morphology-aware losses to ensure that synthesized motions accurately reflect species-specific structures.

\subsection{Unified Human-Animal Motion Modeling}

Building a unified framework for both human and animal motion remains challenging. WalkTheDog~\cite{li2024walkthedog} learns a shared phase manifold for unsupervised semantic and temporal alignment across morphologically different subjects, enabling applications such as motion transfer without requiring explicit skeleton correspondences. AnyTop~\cite{gat2025anytop} generates natural motions aligned with a character’s biomechanical characteristics by conditioning on its skeletal structure. However, the generated motions are sampled from the training distribution and remain relatively uncontrollable. UniMoGen~\cite{khani2025unimogen} synthesizes motions based on the input character’s style, trajectory, and motion history.
In contrast to methods requiring user-specified skeletons, our X-MoGen enables cross-species motion generation solely from free-form text.
\section{Method}
\begin{figure*}[t!]
    \centering
    \includegraphics[width=1.9\columnwidth]{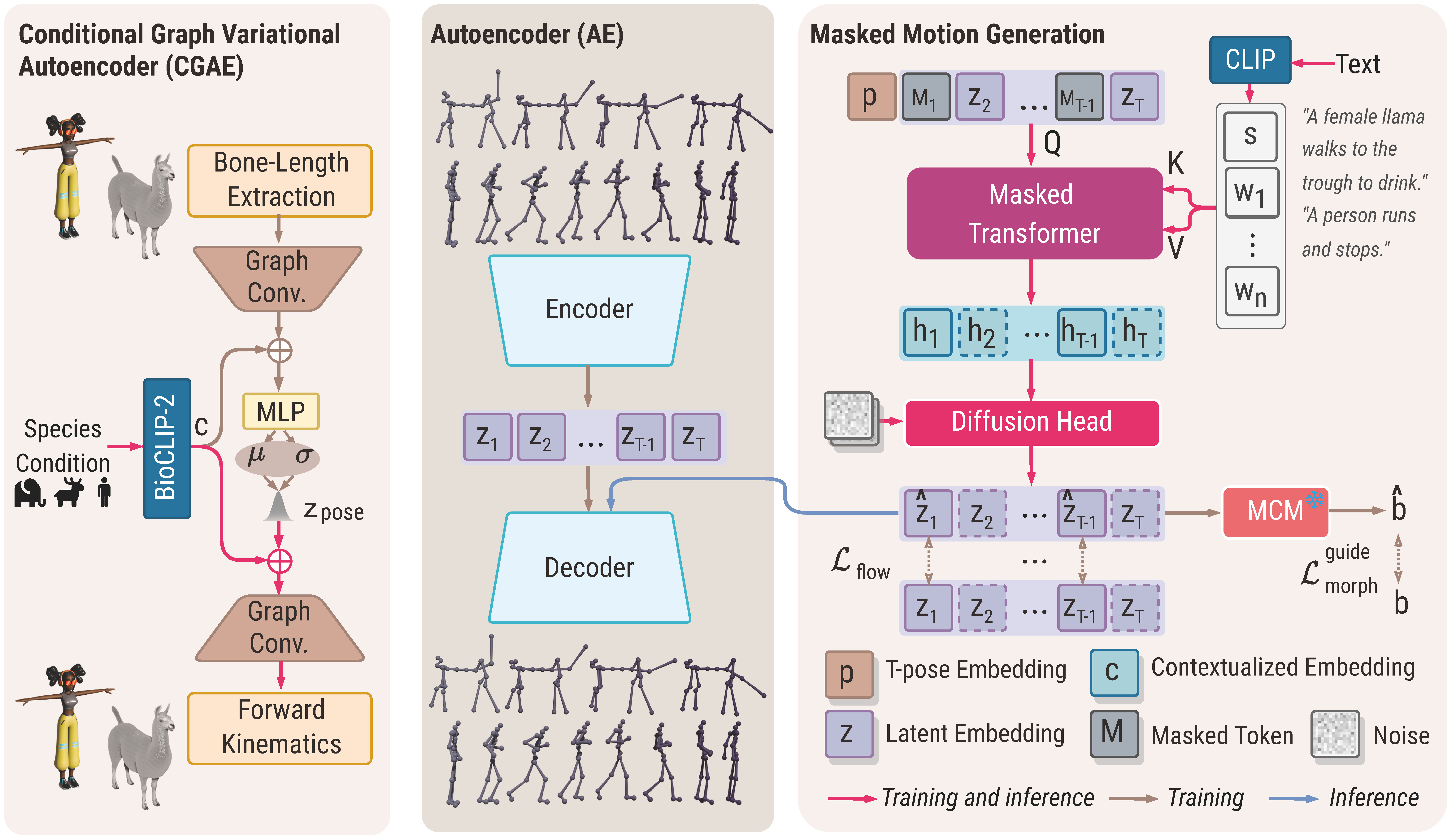}
    \caption{\textbf{Overview of the X-MoGen architecture}. Our two-stage framework first learns a T-pose prior and a compact latent motion space via a Conditional Graph Autoencoder (CGAE) and an Autoencoder (AE). In the second stage, a Masked Transformer (M-Trans) conditions a diffusion model to generate motion from noise, guided by the text description and species-specific T-pose priors produced by the CGAE. An auxiliary MCM promotes morphological constraints in the generated motions.
}
    \label{fig:arch}
\end{figure*}

We address the challenge of cross-species motion generation using X-MoGen, a novel two-stage framework illustrated in Figure~\ref{fig:arch}.
In the first stage, a Conditional Graph Variational Autoencoder (CGAE) generates canonical T-pose priors conditioned on species, while an Autoencoder (AE) encodes motion sequences into a compact latent space.
In the second stage, a Masked Transformer (M-Trans) synthesizes motion from text, conditioned on T-pose priors derived from species cues in the input. The output is further refined by a diffusion-based generation head.
To ensure anatomical plausibility across species, a Morphological Consistency Module (MCM) enforces structural coherence.

\subsection{Stage 1: Feature Modeling}

\paragraph{T-pose Modeling via CGAE.}
We propose a Conditional Graph Variational Autoencoder (CGAE) to model species-specific skeletal morphology and generalize to unseen species. CGAE integrates a conditional variational framework~\cite{sohn2015learning} with graph convolutional encoders and decoders~\cite{kipf2016semi}, explicitly treating the skeletal structure as a graph.
CGAE is trained on bone-length vectors $\mathbf{b} \in \mathbb{R}^{N_b}$ extracted from T-poses using \textit{Bone-Length Extraction}. It learns a latent distribution conditioned on species-level semantic embeddings $\mathbf{c}$ derived from BioCLIP-2's text encoder~\cite{gu2025bioclip}. Both encoder and decoder operate on a fixed-topology graph, with bone lengths as node features.

The training objective balances reconstruction accuracy and latent regularization using the evidence lower bound (ELBO)~\cite{kingma2013auto}:
\begin{equation}
\label{eq:cvae_loss}
\mathcal{L}_{\text{CGAE}} = \left\| \hat{\mathbf{b}} - \mathbf{b} \right\|_2^2 + \beta \, D_{\text{KL}}\left( q_\phi(\mathbf{z}_{\text{pose}} \mid \mathbf{b}, \mathbf{c}) \,\|\, p(\mathbf{z}_{\text{pose}}) \right),
\end{equation}
where $\hat{\mathbf{b}}$ is the reconstructed bone-length vector, $p(\mathbf{z}_{\text{pose}}) = \mathcal{N}(\mathbf{0}, \mathbf{I})$ is a standard Gaussian prior, $D_{\text{KL}}(\cdot \,\|\, \cdot)$ denotes the Kullback–Leibler divergence that regularizes the posterior, and $\beta$ is a weighting factor controlling the trade-off between reconstruction fidelity and latent smoothness.

Once trained, CGAE can generate canonical T-poses for arbitrary quadruped species. Given a condition vector $\mathbf{c}$ and a latent sample $\mathbf{z}_{\text{pose}} \sim \mathcal{N}(\mathbf{0}, \mathbf{I})$, the decoder predicts a bone-length vector $\hat{\mathbf{b}}$. Final joint positions $\mathbf{P}$ are recovered via \textit{Forward Kinematics} based on a predefined kinematic tree~\cite{parent2012computer}.

\paragraph{Motion Representation via AE.}
To encode motion sequences into an efficient and structured representation~\cite{rombach2022high}, we employ an Autoencoder (AE)~\cite{bank2023autoencoders} that operates on raw motion inputs $\mathbf{X} = [\mathbf{x}_1, \ldots, \mathbf{x}_L]$. The encoder maps $\mathbf{X}$ to a temporally downsampled latent sequence $\mathbf{Z} = [\mathbf{z}_1, \ldots, \mathbf{z}_T] \in \mathbb{R}^{T \times d}$, where $T = \lfloor L / 4 \rfloor$ and $d$ denotes the latent dimension. The encoder consists of convolutional layers with temporal strides and residual connections~\cite{he2016deep}, while the decoder uses transposed convolutions to reconstruct the original sequence.

We train the AE by minimizing a weighted combination of reconstruction and morphological loss:
\begin{equation}
\mathcal{L}_{\text{AE}} = \mathcal{L}^{\text{recon}}_{\text{MSE}} + \lambda_{\text{morph}}^{\text{recon}} \cdot \mathcal{L}_{\text{morph}}^{\text{recon}}.
\end{equation}

The reconstruction loss is defined as:
\begin{equation}
\mathcal{L}^{\text{recon}}_{\text{MSE}} = \frac{1}{L} \sum_{t=1}^{L} \left\| \hat{\mathbf{x}}_t - \mathbf{x}_t \right\|_2^2,
\end{equation}
where $\hat{\mathbf{x}}_t$ is the reconstructed motion at time $t$.

The morphological loss enforces structural fidelity by comparing bone lengths computed from motions:
\begin{equation}
\mathcal{L}_{\text{morph}}^{\text{recon}} = \frac{1}{L} \sum_{t=1}^{L} \left\| \mathcal{B}(\hat{\mathbf{x}}_t) - \mathcal{B}(\mathbf{x}_t) \right\|_2^2,
\end{equation}
where $\mathcal{B}(\cdot)$ extracts bone lengths via pairwise joint distances.

\subsection{Stage 2: Masked Motion Generation}
Recent studies have demonstrated the effectiveness of masked modeling~\cite{li2023mage,guo2024momask,li2024autoregressive,meng2025rethinking}. Building on this, we adopt a Masked Transformer (M-Trans) with a diffusion head to synthesize motion sequences. The model learns to reconstruct masked motion tokens through a denoising process, guided by text description.

\paragraph{Masked Motion Modeling.}
M-Trans serves as a context-aware backbone that encodes semantically rich representations for motion generation. Given a latent motion sequence $\mathbf{Z} \in \mathbb{R}^{T \times d}$, a random subset of temporal positions $\mathcal{M} \subset \{1, \ldots, T\}$ is masked by replacing the corresponding tokens with a learnable masked token \texttt{[M]}.
The canonical T-pose $\mathbf{P}$ is projected to an embedding $\mathbf{p} \in \mathbb{R}^{d}$ via linear transformation and concatenated with the masked sequence as the first token, forming the input sequence. The query tensor is computed as:
\begin{equation}
\mathbf{Q} = \text{Proj}_Q([\mathbf{p}; \mathbf{Z}_{\text{masked}}]).
\end{equation}

Textual guidance is extracted from CLIP's text encoder~\cite{radford2021learning}, which provides a sentence-level feature $\mathbf{s}$ and word-level features $\mathbf{W}$. These features are projected to obtain keys and values for cross-attention:
\begin{equation}
\mathbf{K} = \text{Proj}_K([\mathbf{s}; \mathbf{W}]), \quad
\mathbf{V} = \text{Proj}_V([\mathbf{s}; \mathbf{W}]).
\end{equation}

The Transformer fuses motion and text features~\cite{vaswani2017attention}, producing a contextualized representation $\mathbf{H} \in \mathbb{R}^{(T+1) \times d}$:
\begin{equation}
\mathbf{H} = \text{Transformer}(\mathbf{Q}, \mathbf{K}, \mathbf{V}).
\end{equation}

\paragraph{Diffusion-based Completion Head.}
The contextualized representation~$\mathbf{H}$ from the M-Trans backbone serves as the conditioning signal for a diffusion-based completion head. This head is realized by a set of Multi-Layer Perceptrons (MLPs), which function as the velocity network~$\mathbf{v}_\theta$. For each masked position~$i \in \mathcal{M}$, its corresponding contextualized embedding~$\mathbf{h}_i$ is extracted from~$\mathbf{H}$. The velocity network is optimized to predict the deterministic velocity field that maps noise~$\hat{\mathbf{z}}_i$ to the ground-truth data~$\mathbf{z}_i$, following the flow-matching paradigm introduced by SiT~\cite{ma2024sit}. The optimization uses the following loss, conditioned on~$\mathbf{h}_i$:
\begin{equation}
\label{eq:flow_loss}
\mathcal{L}_{\text{flow}} = \frac{1}{|\mathcal{M}|} \sum_{i \in \mathcal{M}} \mathbb{E}_{\tau, \hat{\mathbf{z}}_i, \mathbf{z}_i} \left[ \|\mathbf{v}_\theta(\mathbf{z}_{\tau,i}, \tau, \mathbf{h}_i) - (\mathbf{z}_i - \hat{\mathbf{z}}_i)\|_2^2 \right],
\end{equation}
where $\mathbf{z}_{\tau,i} = (1-\tau)\hat{\mathbf{z}}_i + \tau\mathbf{z}_i$ is the interpolated latent state at time~$\tau$.

\paragraph{Morphological Consistency Module.}
To encourage realistic and consistent skeletal structures in generated motions, we design the Morphological Consistency Module (MCM), denoted as $f_{\text{MCM}}$. This module serves as a morphology-aware regularizer that maps the latent motion sequence $\hat{\mathbf{Z}}$ to a static vector representing the skeleton's bone lengths. Specifically, a Gated Recurrent Unit (GRU)~\cite{chung2014empirical} encodes the temporal dynamics of $\hat{\mathbf{Z}}$, and its final hidden state is processed by a MLP to predict the bone-length vector. 
The MCM is pre-trained on the latent space of AE and remains frozen during motion generation. To ensure morphological consistency, we introduce a guidance loss that penalizes deviations from a canonical bone-length vector $\mathbf{b}$, computed from the ground-truth bone lengths:
\begin{equation}
\label{eq:morph_loss}
\mathcal{L}_{\text{morph}}^{\text{guide}} = \left\| f_{\text{MCM}}(\hat{\mathbf{Z}}) - \mathbf{b} \right\|_2^2.
\end{equation}
Gradients from this loss are backpropagated only through the masked positions $\mathcal{M}$ in $\hat{\mathbf{Z}}$, while unmasked positions are detached.

\paragraph{Training Objective.}
The total training loss combines the flow objective and morphological consistency:
\begin{equation}
\mathcal{L}_{\text{gen}} = \mathcal{L}_{\text{flow}} + \lambda_{\text{morph}}^{\text{guide}} \cdot \mathcal{L}_{\text{morph}}^{\text{guide}}.
\label{eq:total_loss}
\end{equation}

\subsection{Inference}
During inference, the latent sequence $\mathbf{Z}^{(0)} \in \mathbb{R}^{T \times d}$ is initialized with a learnable masked token \texttt{[M]} at all positions, where the superscript $(0)$ denotes the initial iteration step. The full sequence is generated over $R$ iterative steps.
At each iteration step $r$, a subset of the masked positions is selected for refinement. For these positions, latent embeddings are generated via a diffusion-based denoising process. Specifically, Gaussian noise $\hat{\mathbf{z}}_i$ is initialized for each selected position $i$, and $N$ denoising steps are performed using the diffusion head, conditioned on the text prompt and the unmasked tokens generated so far.

To improve text alignment, classifier-free guidance~\cite{ho2022classifier} is applied at each denoising step to compute the guided velocity:
\begin{equation}
\mathbf{v}_{\text{guided},i}^{(n)} = \mathbf{v}_{\text{uncond},i}^{(n)} + \omega \cdot \left(\mathbf{v}_{\text{cond},i}^{(n)} - \mathbf{v}_{\text{uncond},i}^{(n)}\right),
\end{equation}
where $\omega > 0$ is the guidance scale, and the superscript $(n)$ denotes the $n$-th denoising step within each iteration.

After denoising, the generated latent embeddings $\mathbf{z}_i$ replace the corresponding \texttt{[M]} positions in $\mathbf{Z}^{(r)}$, yielding the updated sequence $\mathbf{Z}^{(r+1)}$. This process repeats until all positions are filled. Finally, the completed latent sequence $\mathbf{Z}$ is decoded by the AE to produce the motion output $\mathbf{X}$.

\section{Experiments}

\begin{figure*}[t!]
    \centering
    \includegraphics[width=2.0\columnwidth]{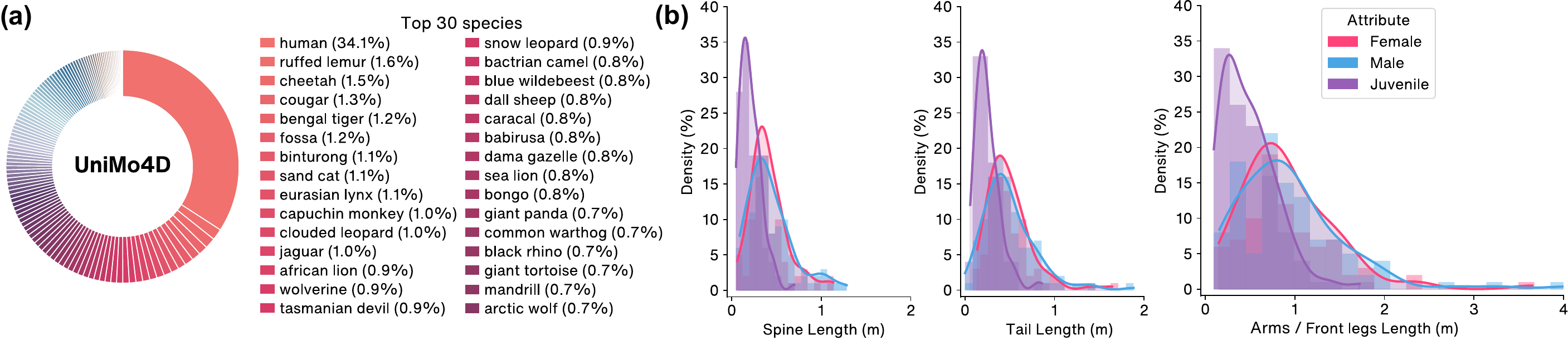}
    \caption{\textbf{Statistics of the UniMo4D dataset.} (a) Species distribution. (b) Length distribution of key bones.}
    \label{fig:dataset}
\end{figure*}
\begin{table*}[ht]
\centering
\vspace{-7pt}
\renewcommand{\arraystretch}{1.2}
\small
\begin{tabular}{llccccccc}
\toprule
\multirow{2}{*}{Methods} & \multirow{2}{*}{Venue} & \multicolumn{3}{c}{R-Precision $\uparrow$} & \multirow{2}{*}{FID $\downarrow$} & \multirow{2}{*}{MM-Dist $\downarrow$} & \multirow{2}{*}{Diversity $\rightarrow$} & \multirow{2}{*}{MME $\downarrow$} \\
\cmidrule(lr){3-5}
& & Top-1 $\uparrow$ & Top-2 $\uparrow$ & Top-3 $\uparrow$ & & & & \\
\midrule
Real & $-$ & $0.899^{\pm .001}$ & $0.972^{\pm .001}$ & $0.989^{\pm .001}$ & $0.000^{\pm .000}$ & $0.470^{\pm .133}$ & $19.407^{\pm .255}$ & - \\
\midrule
MDM & ICLR'23 & $0.283^{\pm .007}$ & $0.444^{\pm .010}$ & $0.556^{\pm .010}$ & $4.684^{\pm .833}$ & $7.766^{\pm .109}$ & $18.199^{\pm .250}$ & $0.313^{\pm .000}$ \\
T2M-GPT & CVPR'23 & $0.382^{\pm .002}$ & $0.463^{\pm .002}$ & $0.677^{\pm .002}$ & $5.058^{\pm .059}$ & $5.196^{\pm .014}$ & $18.452^{\pm .316}$ & $0.409^{\pm .000}$ \\
AttT2M & ICCV'23 & $0.334^{\pm .001}$ & $0.466^{\pm .001}$ & $0.695^{\pm .001}$ & $5.895^{\pm .192}$ & $5.695^{\pm .030}$ & $15.895^{\pm .219}$ & $0.456^{\pm .000}$ \\
MMM & CVPR'24 & $0.585^{\pm .001}$ & $0.746^{\pm .001}$ & $0.822^{\pm .001}$ & $5.200^{\pm .027}$ & $3.742^{\pm .004}$ & $18.225^{\pm .284}$ & $0.390^{\pm .000}$ \\
MoMask & CVPR'24 & $0.554^{\pm .002}$ & $0.752^{\pm .002}$ & $0.845^{\pm .001}$ & $0.745^{\pm .018}$ & $3.602^{\pm .006}$ & $19.151^{\pm .285}$ & $0.220^{\pm .000}$ \\
MARDM & CVPR'25 & $0.823^{\pm .001}$ & $0.927^{\pm .001}$ & $0.960^{\pm .001}$ & $0.189^{\pm .002}$ & $0.874^{\pm .001}$ & $18.706^{\pm .423}$ & $0.251^{\pm .000}$ \\
AniMo & CVPR'25 & $0.596^{\pm .003}$ & $0.786^{\pm .002}$ & $0.869^{\pm .002}$ & $0.533^{\pm .007}$ & $3.174^{\pm .009}$ & $17.955^{\pm .305}$ & $0.211^{\pm .000}$ \\
\textbf{X-MoGen} & Ours & $\textbf{0.848}^{\pm .001}$ & $\textbf{0.941}^{\pm .001}$ & $\textbf{0.968}^{\pm .001}$ & $\textbf{0.050}^{\pm .001}$ & $\textbf{0.742}^{\pm .002}$ & $\textbf{19.332}^{\pm .480}$ & $\textbf{0.201}^{\pm .000}$ \\
\bottomrule
\end{tabular}
\caption{Comparison of text-driven motion generation on the UniMo4D dataset. The right arrow (→) indicates that values closer to the ground truth motion are better. The best results for each metric are shown in bold.}
\label{tab:results}
\vspace{-6pt}
\end{table*}

\begin{table*}[ht]
\centering
\vspace{-7pt}
\renewcommand{\arraystretch}{1.2} %
\small 
\begin{tabular}{llccccccc}
\toprule
\multirow{2}{*}{Methods} & \multirow{2}{*}{Venue} & \multicolumn{3}{c}{R-Precision $\uparrow$} & \multirow{2}{*}{FID $\downarrow$} & \multirow{2}{*}{MM-Dist $\downarrow$} & \multirow{2}{*}{Diversity $\rightarrow$} & \multirow{2}{*}{MME $\downarrow$} \\
\cmidrule(lr){3-5}
& & Top-1 $\uparrow$ & Top-2 $\uparrow$ & Top-3 $\uparrow$ & & & & \\
\midrule
Real & $-$ & $0.594^{\pm .003}$ & $0.826^{\pm .003}$ & $0.924^{\pm .002}$ & $0.000^{\pm .000}$ & $0.310^{\pm .001}$ & $20.015^{\pm .399}$ & - \\
\midrule
MDM & ICLR'23 & $0.099^{\pm .003}$ & $0.186^{\pm .003}$ & $0.268^{\pm .004}$ & $28.481^{\pm 1.678}$ & $11.490^{\pm .076}$ & $23.587^{\pm .652}$ & $0.310^{\pm .002}$ \\
T2M-GPT & CVPR'23 & $0.101^{\pm .001}$ & $0.184^{\pm .001}$ & $0.256^{\pm .001}$ & $55.775^{\pm .223}$ & $11.025^{\pm .022}$ & $13.729^{\pm .267}$ & $0.457^{\pm .000}$ \\
AttT2M & ICCV'23 & $0.093^{\pm .001}$ & $0.180^{\pm .001}$ & $0.254^{\pm .002}$ & $65.045^{\pm 1.172}$ & $15.279^{\pm .051}$ & $16
.125^{\pm .252}$ & $0.485^{\pm .000}$ \\
MMM & CVPR'24 & $0.113^{\pm .003}$ & $0.213^{\pm .001}$ & $0.298^{\pm .002}$ & $49.840^{\pm .242}$ & $10.489^{\pm .025}$ & $13.887^{\pm .227}$ & $0.445^{\pm .000}$ \\
MoMask & CVPR'24 & $0.138^{\pm .002}$ & $0.261^{\pm .003}$ & $0.369^{\pm .003}$ & $24.441^{\pm .193}$ & $8.180^{\pm .020}$ & $\textbf{18.383}^{\pm .248}$ & $0.235^{\pm .000}$ \\
MARDM & CVPR'25 & $0.143^{\pm .002}$ & $0.265^{\pm .001}$ & $\textbf{0.374}^{\pm .002}$ & $34.124^{\pm .250}$ & $\textbf{7.690}^{\pm .004}$ & $13.779^{\pm .305}$ & $0.301^{\pm .000}$ \\
AniMo & CVPR'25 & $0.114^{\pm .002}$ & $0.213^{\pm .002}$ & $0.304^{\pm .002}$ & $45.605^{\pm .245}$ & $9.688^{\pm .016}$ & $15.594^{\pm .353}$ & $0.238^{\pm .000}$ \\
\textbf{X-MoGen} & Ours & $\textbf{0.148}^{\pm .002}$ & $\textbf{0.271}^{\pm .002}$ & $0.372^{\pm .002}$ & $\textbf{19.935}^{\pm .093}$ & $7.939^{\pm .010}$ & $17.652^{\pm .230}$ & $\textbf{0.229}^{\pm .000}$  \\
\bottomrule
\end{tabular}
\caption{Comparison of text-driven motion generation on the UniMo4D unseen species test dataset. The right arrow (→) indicates that values closer to the ground truth motion are better. The best results for each metric are shown in bold.}
\label{tab:results_ood} 
\vspace{-6pt}
\end{table*}

\paragraph{Dataset.}

To support joint modeling of human and animal motions, we construct UniMo4D, a unified dataset that integrates HumanML3D~\cite{t2m}, KIT-ML~\cite{Plappert2016}, and AniMo4D~\cite{wang2025animo}. Among various motion representations~\cite{t2m,lu2023humantomato,meng2025rethinking,meng2025absolute,xiao2025motionstreamer}, we adopt the \textit{essential feature groups}~\cite{meng2025rethinking}, a compact and effective scheme. Each frame is represented as a vector of dimension $N_j \times 3 + 1$, where $N_j$ denotes the number of joints. The root joint features include rotation velocity, linear velocity, and height (y-axis), while the remaining joints are represented by 3D positions.
To create a unified representation, we first standardize the skeletal structures from all datasets. Specifically, we remap joints from KIT-ML and reorder those from AniMo4D to align with the HumanML3D convention~\cite{petrovich23tmr,chen2023unimocap}, scaling bone lengths accordingly. All skeletons are then standardized to a common 25-joint topology, which incorporates three virtual tail joints to accommodate diverse morphologies. For tailless species (\textit{e.g.}, humans), these virtual joints are initialized to be co-located with the pelvic root. This neutral configuration ensures a consistent data structure and prevents motion artifacts, without compromising the fidelity of core body movements~\cite{yang2023unifiedgesture}. As a result, every motion sequence in UniMo4D is represented by a skeleton with $N_j = 25$ joints and a consistent topology.

Motion sequences with lengths $L \in (18, 300)$ are retained, resulting in 118,663 sequences spanning 115 species. Figure~\ref{fig:dataset} (a) presents the species distribution, and Figure~\ref{fig:dataset} (b) illustrates bone length variations across species, highlighting the challenge of unified modeling across diverse morphologies.

In our experiments, five species comprising 5,179 motion sequences are held out as an unseen-species test set. The remaining species are split into training, validation, and test sets with proportions of 0.80, 0.05, and 0.15, respectively.

\paragraph{Evaluation Metrics.}

A text-motion matching model is retrained on the UniMo4D dataset to serve as the evaluation backbone~\cite{t2m,meng2025rethinking}. Based on its learned feature space, motion quality, semantic alignment, diversity, and morphological consistency are evaluated.
\textbf{Frechet Inception Distance (FID)} assesses motion realism by comparing the feature distributions of generated and real motions.  
\textbf{R-Precision} measures text-motion alignment through retrieval accuracy, while \textbf{Multimodal Distance (MM-Dist)} quantifies the average distance between motion and text embeddings.  
\textbf{Diversity} measures variability across motions generated from different prompts.

To evaluate morphological plausibility, \textbf{Mean Morphological Error (MME)} is introduced to measure skeletal consistency over time. Specifically, it quantifies deviations in bone lengths from a reference skeleton, where $\hat{\mathbf{b}}_t \in \mathbb{R}^{N_b}$ denotes the bone lengths at frame $t$, and $\mathbf{b} \in \mathbb{R}^{N_b}$ is the corresponding reference vector. MME is defined as the average L1 distance over the sequence:
\begin{equation}
\label{eq:mme}
\text{MME} = \frac{1}{L \cdot N_b} \sum_{t=1}^{L} \|\hat{\mathbf{b}}_t - \mathbf{b}\|_1
\end{equation}
where $L$ is the sequence length and $N_b$ is the number of bones. Lower MME reflects more consistent species-specific morphology, reducing unrealistic bone deformations.

\paragraph{Baselines.}
To the best of our knowledge, X-MoGen is the first unified framework designed for cross-species motion generation involving both humans and animals. As no existing method is specifically designed for this setting, we compare our approach against recent state-of-the-art models developed for either human or animal motion generation.
Specifically, we evaluate several human motion baselines, including MDM~\cite{tevet2023human}, T2M-GPT~\cite{zhang2023generating}, AttT2M~\cite{zhong2023attt2m}, MMM~\cite{pinyoanuntapong2024mmm}, MoMask~\cite{guo2024momask}, and MARDM~\cite{meng2025rethinking}, along with AniMo~\cite{wang2025animo}, which focuses on animal motion.
All baseline models are retrained from scratch on the UniMo4D dataset using their official implementations and original data formats~\cite{xiao2025motionstreamer,meng2025absolute}. Although these methods use different internal data formats, we extract essential motion features from their outputs to enable a fair and consistent evaluation~\cite{meng2025rethinking}.

\subsection{Quantitative Results}

We evaluate X-MoGen on the UniMo4D dataset. All experiments are repeated 10 times, and we report mean values with 95\% confidence intervals~\cite{guo2024momask,wang2025animo}.
Table~\ref{tab:results} shows the results on the test set. X-MoGen outperforms all baselines with significantly lower FID and higher R-Precision, highlighting the superior motion quality and text alignment of our method.
Table~\ref{tab:results_ood} reports performance on the unseen species test set, which includes species excluded during training and presents a more challenging generalization scenario. X-MoGen maintains strong performance across Top-1, Top-2, FID, and MME, which are the primary metrics for evaluating motion plausibility and morphological consistency.
Table~\ref{tab:results_ae} compares several representative motion compression methods for reconstruction. Among quantization-based approaches, VQ~\cite{van2017neural}, FSQ~\cite{mentzer2023finite}, and RVQ~\cite{lee2022autoregressive,zeghidour2021soundstream} all struggle to capture the complexity of the UniMo4D dataset, although FSQ and RVQ offer moderate improvements over vanilla VQ. Our AE does not use quantization and achieves much better reconstruction accuracy. This shows that continuous latent modeling works better for diverse motion data.

\begin{figure*}[t!]
    \centering
    \includegraphics[width=1.8\columnwidth]{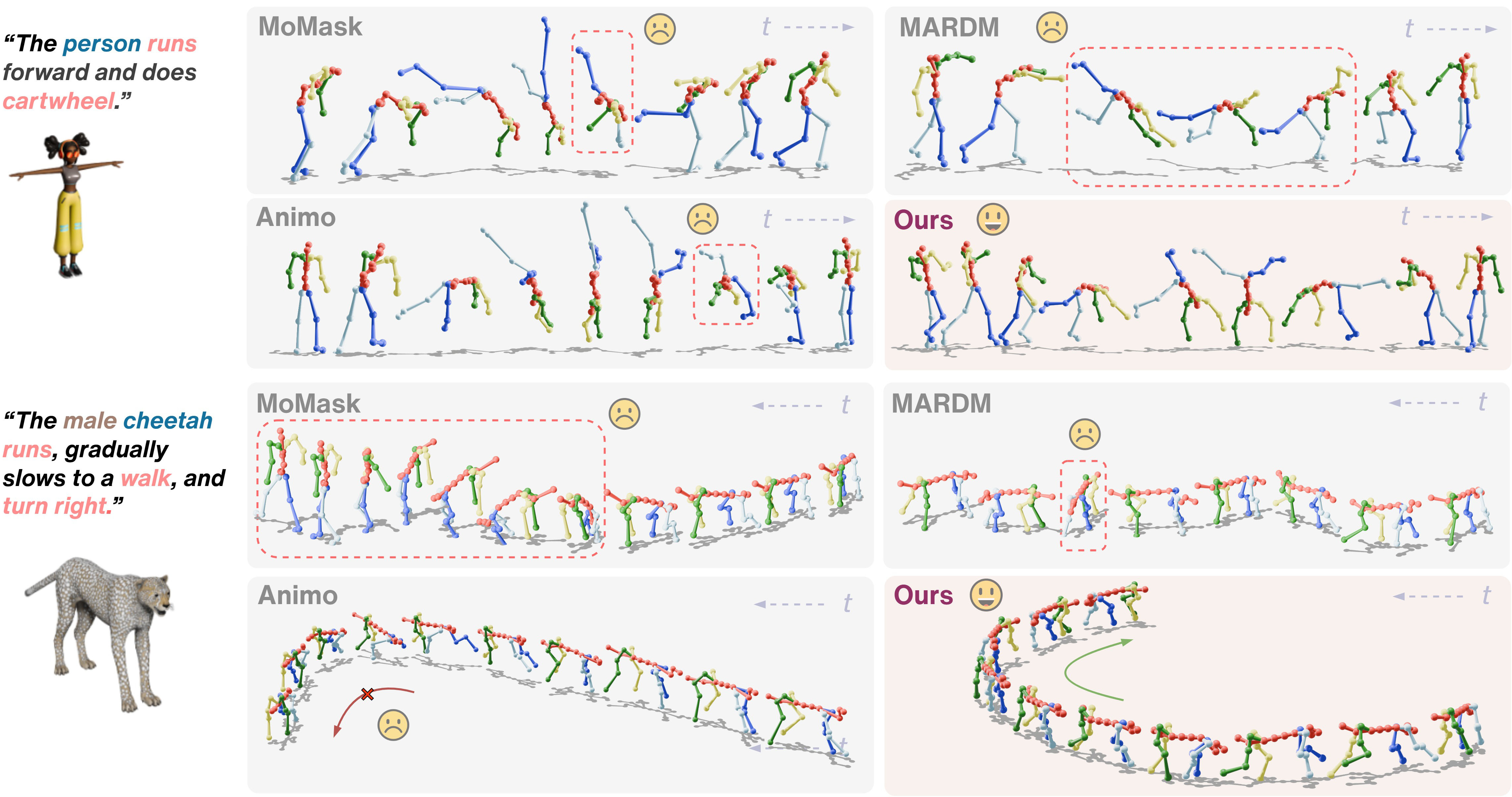}
    \caption{Qualitative results for human and unseen animal motion. Red dashed boxes and red arrows highlight implausible motion artifacts.}
    \label{fig:case}
\end{figure*}

\subsection{Qualitative Results}

To further assess the effectiveness of our model, we qualitatively compare X-MoGen with three strong baselines: MoMask, MARDM, and AniMo, selected based on their quantitative performance. Figure~\ref{fig:case} presents two representative examples.
In the human motion case, X-MoGen accurately captures the prompt containing ``running'' and ``cartwheel'', generating a coherent sequence with a smooth transition into the cartwheel. In contrast, MoMask reverses the intended action order, while MARDM and AniMo produce only the cartwheel, omitting the preceding context. Notably, the virtual tail nodes introduced for topological consistency do not degrade motion realism across all methods.
In the unseen species case, X-MoGen generates motion that is both morphologically consistent and semantically faithful to the prompt. MoMask fails by abruptly morphing the animal into a human, MARDM misinterprets the prompt and produces unstable body rotations, and AniMo incorrectly generates a left-turn motion.
These results demonstrate that X-MoGen effectively unifies human and animal motion synthesis, exhibiting strong generalization across species. 

\subsection{Ablation Study}

We conduct ablation experiments to evaluate the contribution of each component. The ablated elements include: (1) word-level features $\mathbf{W}$ from CLIP; (2) the canonical T-pose prior from CGAE; (3) the morphological reconstruction loss $\mathcal{L}{\text{morph}}^{\text{recon}}$ in the AE; and (4) the morphological guidance loss $\mathcal{L}{\text{morph}}^{\text{guide}}$ in the MCM.

As shown in Table~\ref{tab:abla}, removing word-level embeddings slightly increases FID, indicating that fine-grained text features enhance semantic alignment. Excluding the CGAE prior raises both MME and FID, highlighting its role in producing structurally consistent T-pose priors. While removing MCM leads to a marginal increase in FID (+0.001), it notably worsens MME, suggesting its importance for motion consistency. Similarly, ablating the AE reconstruction loss also results in a performance drop.

These results validate the necessity of each component and demonstrate their complementary roles in achieving high-quality, semantically aligned motion generation.

\begin{table}[ht]
\centering
\small 
\renewcommand{\arraystretch}{1.2} 
\begin{tabular}{lcccc}
\toprule
Compressor & Top-1 $\uparrow$ & FID $\downarrow$ & MM-Dist $\downarrow$ & MME $\downarrow$ \\
\midrule
VQ (T2M-GPT) & $0.581$ & $5.010$ & $4.015$ & $0.389$ \\
FSQ (MMM) & $0.629$ & $4.272$& $3.414$ & $0.388$ \\
RVQ (MoMask) & $0.700$ & $0.451$& $2.299$ & $0.220$ \\
AE (X-MoGen)& $\textbf{0.898}$ & $\textbf{0.008}$ & $\textbf{0.496}$ & $\textbf{0.064}$ \\
\bottomrule
\end{tabular}
\caption{Quantitative results of different motion compressors on the UniMo4D test set.}
\label{tab:results_ae}
\end{table}

\begin{table}[ht]
\centering
\renewcommand{\arraystretch}{1.2} 
\small 
\renewcommand{\arraystretch}{1.2} 
\begin{tabular}{lcccc} 
\toprule
Methods & Top-1$ \uparrow$ & FID $\downarrow$ & MM-Dist $\downarrow$ & MME $\downarrow$ \\ 
\midrule
X-MoGen & $\textbf{0.848}$ & $0.050$ & $\textbf{0.742}$ & $\textbf{0.201}$ \\
\textit{w/o} $\mathbf{W}$ & $0.835$ & $0.053$ & $0.806$ & $0.205$ \\
\textit{w/o} CGAE & $0.841$ & $0.055$ & $0.758$ & $0.228$ \\
\textit{w/o} $\mathcal{L}_{\text{morph}}^{\text{recon}}$ & $0.843$ & $0.108$ & $0.845$ & $0.237$ \\
 \textit{w/o} MCM & $\textbf{0.848}$ & $\textbf{0.049}$ & $0.743$ & $0.238$ \\
\bottomrule
\end{tabular}
\caption{Ablation study on the UniMo4D test set.} 
\label{tab:abla}
\end{table}

\subsection{Applications}

\paragraph{Universal Motion Generation.}
X-MoGen presents a unified framework for generating both human and animal motions from text within a shared representation space. The framework supports humans and 114 animal species, leveraging CGAE-generated, species-specific T-poses to generalize to new quadrupeds without additional training. This enables intuitive AI-driven motion creation and supports diverse applications including content production, biology-inspired simulation, digital storytelling, and interactive media~\cite{abootorabi2025generative}.

\paragraph{Cross-Species Motion Transformation.}
X-MoGen enables smooth motion transitions between species. Given two motion sequences (e.g., a tiger walking and a human walking), the AE encoder extracts latent features, and mask tokens \texttt{[M]} are inserted at the transition point. M-Trans, with the diffusion head and AE decoder, generates realistic cross-species transitions. As shown in Figure~\ref{fig:teaser}, this enables smooth morphing between animal and human motions, supporting seamless cross-species motion editing and completion, valuable for animation and automated shapeshifting in film~\cite{akber2023deep,tessler2024maskedmimic}.

\section{Conclusion}
We introduced \textbf{X-MoGen}, a unified framework for generating motions across both human and animal domains. X-MoGen models skeletal structures with a CGAE and compresses motions into a latent space with an AE. It further incorporates masked motion modeling conditioned on text and a T-pose prior derived from a CGAE. During training, we leverage MCM to encourage morphologically plausible generation. Additionally, we constructed the \textbf{UniMo4D} dataset, which standardizes human and animal motion representations, facilitating future research in cross-species motion generation. Extensive experiments demonstrate that X-MoGen outperforms strong baselines, generating realistic motions with generalization to unseen species.

\section*{Acknowledgments}

This work was supported by the Zhejiang Provincial Natural Science Foundation of China (No. LZ24F030005), the National Natural Science Foundation of China (No. 62576308), the Fundamental Research Funds for the Central Universities (No. 226-2025-00167), and the Scientific Research Foundation of Sichuan Provincial Department of Science and Technology, China (No. 2024YFHZ0001).

\bibliography{aaai2026}
\end{document}